\documentclass[10pt,twocolumn,letterpaper]{article}

\usepackage{cvpr}              

\usepackage{multirow}

\definecolor{cvprblue}{rgb}{0.21,0.49,0.74}
\usepackage[pagebackref,breaklinks,colorlinks,allcolors=cvprblue]{hyperref}

\def\paperID{*****} 
\def\confName{CVPR}
\def\confYear{2026}

\title{Two-Stage Multimodal Framework for Emotion Mimicry Intensity Prediction}

\author{
Dinithi Dissanayake$^{1}$, Shaveen Silva$^{1,2}$, Ovindu Atukorala$^{1,2}$,\\
Prasanth Sasikumar$^{1}$, Suranga Nanayakkara$^{1}$\\
$^{1}$Augmented Human Lab, National University of Singapore, Singapore\\
$^{2}$University of Moratuwa, Sri Lanka\\
{\tt\small \{dinithi, shaveen, ovindu, prasanth, suranga\}@ahlab.org}
}

\begin{document}
\maketitle
\begin{abstract}
We present our submission to the Hume-ABAW10 Emotional Mimicry Intensity (EMI) Challenge, which aims to predict six continuous emotion intensity dimensions: \textit{Admiration}, \textit{Amusement}, \textit{Determination}, \textit{Empathic Pain}, \textit{Excitement}, and \textit{Joy}, from in-the-wild multimodal video clips. We propose a staged multimodal framework that combines textual, acoustic, and visual representations, with an optional motion branch. Our approach first trains modality-specific encoders independently and then fuses their learned representations through a lightweight regressor with modality dropout and controlled encoder adaptation. Across our submitted systems, the best validation performance is obtained by the text--audio--vision--motion fusion model under the expanded 4:1 split, achieving an average Pearson correlation of 0.4722. Although the motion branch yields only very slight gains, its behavior can be interesting to study. Our team was placed third in the EMI challenge, achieving an average Pearson correlation of 0.57 for the test set. Overall, we provide a practical and reproducible baseline for EMI prediction.
\end{abstract}    
\section{Introduction}

Emotion mimicry intensity prediction is an important problem at the intersection of affective computing, multimodal machine learning, and human behaviour understanding. Unlike categorical emotion recognition, the EMI task requires predicting continuous intensity values across multiple affective dimensions, making it particularly sensitive to ambiguity, sparsity, and differences in how emotion is expressed across clips. The Hume-ABAW10 EMI Challenge further amplifies these difficulties by presenting in-the-wild data\cite{kollias2025advancements} with strong temporal variation, partial modality missingness, and highly imbalanced targets.

To address these challenges, we propose a staged multimodal framework that integrates text, audio, and vision, with an optional motion branch. Rather than training a single end-to-end multimodal model from scratch, we first learn modality-specific encoders independently and then combine their representations through a lightweight fusion regressor. This strategy allows each branch to develop a stable unimodal representation before cross-modal interaction, while still permitting limited encoder adaptation during fusion. In addition, modality dropout is used to improve robustness to weak or missing channels.

Our exploratory analysis shows that the EMI dataset exhibits long-tailed sequence lengths, substantial label sparsity, and partial incompleteness in the text modality. These properties directly motivate our design choices. The strong imbalance across target dimensions, especially for \textit{Empathic Pain}, encourages the use of a CCC-oriented regression objective, while the variability in sequence duration motivates temporal pooling and sequence-aware encoders. Missing-text fallback and multimodal fusion are further introduced to reduce reliance on any single modality.

Empirically, we find that text and audio provide the strongest unimodal predictive signals, whereas visual and motion features contribute more selectively to dimensions such as \textit{Joy}, \textit{Amusement}, and \textit{Excitement}. Our final fusion systems consistently outperform the unimodal branches, with the best validation result obtained by the text--audio--vision--motion model on the expanded 4:1 split. Overall, this work makes three contributions: (1) a practical staged training strategy for multimodal EMI prediction, (2) an empirical analysis of how different modalities contribute across emotion axes, and (3) a challenge submission that highlights the complementary role of motion cues in multimodal fusion. We have made our code publicly available at \url{https://github.com/Dinithipurna/MimicMetrics} to support reproducibility.

\section{Related Work}

Significant advancements in affective and behavioural computing have enhanced the understanding and automated detection of emotional expressions, particularly through facial analysis. Contemporary research increasingly demonstrates that leveraging multimodal data, including text, audio, and visual inputs, leads to improved performance across a wide range of affective computing tasks \cite{kollias2023abawvalencearousalestimationexpression}. In particular, multimodal approaches enable models to capture complementary emotional cues that are not present in a single modality~\cite{savchenko2025hsemotion, yu2024efficient}.

Rather than relying solely on raw video inputs, many approaches adopt pretrained encoders to extract high-level, semantically rich representations from each modality. In such methods, features are often learned separately for each modality, such as audio, visual, and textual streams, allowing the model to capture modality-specific characteristics~\cite{hallmen2024unimodal, qiu2024language}. These representations may then be combined through a fusion mechanism, enabling the integration of complementary information across modalities and supporting a more comprehensive understanding of affective signals~\cite{savchenko2025hsemotion, yu2024efficient, yu2025dual}.

Prior work has utilised text encoders such as BERT~\cite{devlin2019bert}, ELECTRA~\cite{clark2020electra}, and GTE-base~\cite{li2023towards} to derive informative textual representations. For the audio modality, pretrained models including VGGish~\cite{hershey2017cnn} and Wav2Vec 2.0~\cite{baevski2020wav2vec} are commonly employed, with some approaches further incorporating handcrafted features such as eGeMAPS~\cite{li2023exploration} to enrich acoustic representations. In the visual domain, architectures such as ResNet~\cite{he2016deep}, EfficientNet~\cite{tan2019efficientnet}, and Masked Autoencoders~\cite{he2022masked} have been adopted to capture expressive facial and scene-level features.

There have been multiple approaches for the fusion of these multimodal features. Ding et al implement a specialised fusion block with cross-attention mechanisms for the integration of multimodal data~\cite{ding2025cross}. Richet et al. \cite{richet2025} proposed a system where each modality is individually processed using TCNs, and subsequently integrated via a co-attention mechanism to perform the final prediction, while Jun Yu et al \cite{yu2025technical} proposed a system with Dual-Stage Cross-Modal Alignment followed by Quality-Aware Dynamic Fusion. 

Motivated by these prior efforts, our submission adopts a two-stage multimodal fusion framework that combines audio, visual, and textual representations, together with an optional motion branch. While the motion branch provides only complementary gains in our current setting, we highlight it as a promising alternative direction for further exploration in emotion mimicry intensity prediction.




\section{Exploratory Data Analysis}

\subsection{Dataset Characteristics}
We first examined dataset composition and input quality across the training and validation splits. The dataset contains 12{,}660 samples in total, comprising 8{,}072 training instances and 4{,}588 validation instances. Sequence durations are highly variable across both splits. In the training set, videos contain $73.33 \pm 62.11$ frames on average (median = 65, range = 1--2{,}255), whereas in the validation set, videos contain $75.81 \pm 191.39$ frames on average (median = 66, range = 1--12{,}500). Although no missing-frame cases were found after frame extraction, modality completeness analysis revealed missing text embeddings for 5.67\% of training samples and 6.34\% of validation samples, while face and audio embeddings were available throughout. Following prior challenge papers, we additionally extracted text embeddings as a complementary semantic modality when available~\cite{savchenko2025hsemotion, yu2025technical, zhang2024affective}. Overall, these results indicate substantial temporal heterogeneity together with partial modality sparsity.

\subsection{Label Distribution}
The target distributions are clearly imbalanced (Tables~\ref{tab:dataset_stats}). Among the emotion dimensions, \textit{Empathic Pain} has the lowest mean intensity (train: 0.0214, validation: 0.0134), whereas \textit{Admiration} and \textit{Excitement} exhibit comparatively higher mean values. Label sparsity is also pronounced, with zero-valued targets dominating several dimensions. For example, \textit{Empathic Pain} contains approximately 89.9\% zero-valued targets in the training split and 92.2\% in the validation split. This strong skew motivated the use of robust regression objectives based on Concordance Correlation Coefficient (CCC), rather than relying solely on mean squared error.

\begin{table*}[t]
\centering
\small
\setlength{\tabcolsep}{4pt}
\begin{tabular}{lrrrrrrrrrrrr}
\toprule
\textbf{Split} & \textbf{Samples} & \textbf{Frames (Mean $\pm$ Std)} & \textbf{Min} & \textbf{Max} & \textbf{Adm.} & \textbf{Amuse.} & \textbf{Determ.} & \textbf{Emp. Pain} & \textbf{Excite.} & \textbf{Joy} \\
\midrule
\textbf{Train} & 8072 & $73.33 \pm 62.11$ & 1 & 2255 & 0.1785 & 0.1253 & 0.0967 & 0.0214 & 0.1420 & 0.1157 \\
\textbf{Valid} & 4588 & $75.81 \pm 191.39$ & 1 & 12500 & 0.1594 & 0.1303 & 0.0668 & 0.0134 & 0.1294 & 0.0994 \\
\bottomrule
\end{tabular}
\caption{Summary statistics for the training and validation splits, including clip length statistics and mean label values for the six emotion dimensions.}
\label{tab:dataset_stats}
\end{table*}

\subsection{Input Variability Across Modalities}
Input variability is substantial across modalities. On the visual side, the sequence-length statistics reveal a pronounced long-tail distribution (Table~\ref{tab:dataset_stats}), with approximately 6.2\%--6.5\% of samples exceeding 120 frames and rare extreme outliers extending beyond 1{,}000 frames. The text modality is less consistent due to missing embeddings in a subset of samples, while face and audio features remain consistently available. These observations motivated a missing-modality fallback strategy together with multimodal fusion, so that the model does not over-rely on any single channel. We note that direct waveform-level noise statistics were not explicitly quantified in this exploratory analysis.

\subsection{Temporal Patterns}
Temporal diagnostics, including the sequence-length distribution and log-scale boxplots, show that the central tendencies of the training and validation splits are similar, with medians around 65--66 frames, but both splits exhibit heavy and non-negligible tails. This suggests that short local context alone may be insufficient, particularly for longer behavioral episodes. Accordingly, our framework incorporates explicit temporal modeling through recurrent encoders and temporal pooling/attention mechanisms prior to multimodal fusion, enabling it to capture both short- and long-range dependencies.

\subsection{Key Takeaways}
Overall, the exploratory analysis reveals three dominant properties of the dataset: (1) substantial temporal variability with long-tail outliers, (2) strong label sparsity and imbalance across the EMI dimensions, and (3) partial incompleteness in the text modality. These findings directly informed our pipeline design, including CCC-oriented optimization, missing-text fallback handling, temporal encoders, and multimodal fusion. In addition, split-shift analyses (Table~\ref{tab:label_dist_shift}, KS test) indicate significant distribution differences between training and validation for several labels, including \textit{Admiration}, \textit{Determination}, \textit{Excitement}, and \textit{Joy}, further highlighting the importance of robust generalization.

\begin{table}[h]
\centering
\scriptsize
\setlength{\tabcolsep}{3pt}
\begin{tabular}{lcccccc}
\toprule
\textbf{Label} & \textbf{Train} & \textbf{Valid} & \textbf{$\Delta\mu$} & \textbf{Zero \%} & \textbf{KS} & \textbf{$p$} \\
\midrule
Adm.  & $0.178{\scriptscriptstyle \pm 0.250}$ & $0.159{\scriptscriptstyle \pm 0.243}$ &  $0.019$ & 55.3 / 60.0 & 0.047 & $<.001$ \\
Amus. & $0.125{\scriptscriptstyle \pm 0.218}$ & $0.130{\scriptscriptstyle \pm 0.224}$ & $-0.005$ & 63.6 / 63.6 & 0.016 & 0.457 \\
Det.  & $0.097{\scriptscriptstyle \pm 0.188}$ & $0.067{\scriptscriptstyle \pm 0.155}$ &  $0.030$ & 67.5 / 74.9 & 0.075 & $<.001$ \\
Emp.  & $0.021{\scriptscriptstyle \pm 0.096}$ & $0.013{\scriptscriptstyle \pm 0.070}$ &  $0.008$ & 89.9 / 92.2 & 0.022 & 0.108 \\
Exc.  & $0.142{\scriptscriptstyle \pm 0.228}$ & $0.129{\scriptscriptstyle \pm 0.218}$ &  $0.013$ & 61.8 / 64.8 & 0.029 & 0.013 \\
Joy   & $0.116{\scriptscriptstyle \pm 0.213}$ & $0.099{\scriptscriptstyle \pm 0.204}$ &  $0.016$ & 67.5 / 72.3 & 0.048 & $<.001$ \\
\bottomrule
\end{tabular}
\caption{Label distribution differences between the training and validation splits. Values are reported as mean $\pm$ std. Zero $\%$ is shown as train/valid.}
\label{tab:label_dist_shift}
\end{table}

\section{Method}

\subsection{Overview}

\begin{figure*}
    \centering
    \includegraphics[width=0.95\linewidth]{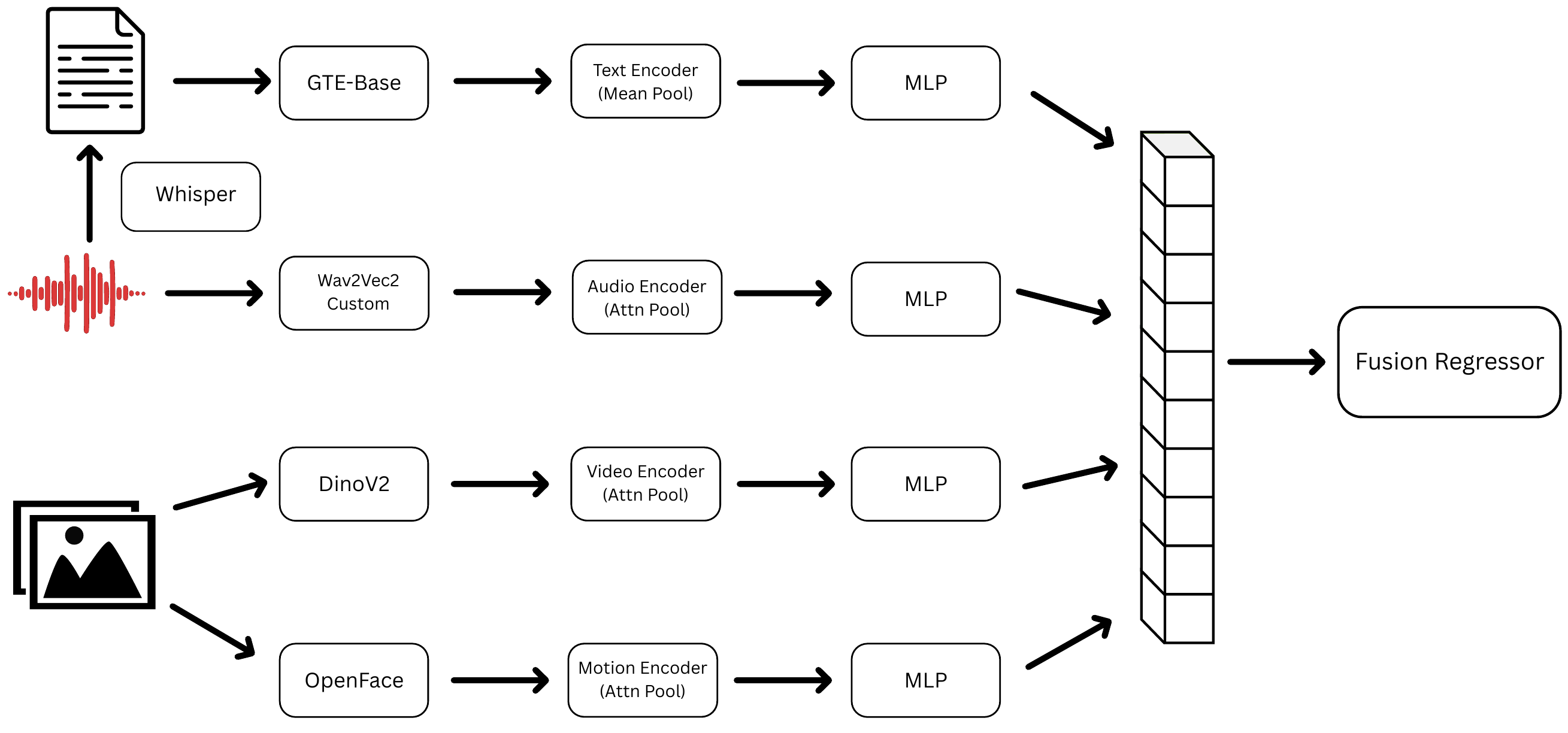}
    \caption{Illustration of our two-stage multimodal framework for Emotional Mimicry Intensity (EMI) prediction. The framework combines text, audio, and visual representations, with an optional motion branch as an exploratory extension. In the first stage, unimodal encoders are trained independently to learn stable modality-specific representations. In the second stage, the learned embeddings are fused through a lightweight regression head to predict six continuous emotion intensity dimensions. Text features are encoded with an MLP-based encoder, while the audio, visual, and motion branches use attention-based temporal pooling to summarize variable-length sequences. Modality dropout and limited encoder adaptation are applied during fusion training to improve robustness and support cross-modal integration.}
    \label{fig:main}
\end{figure*}

We propose a two-stage multimodal framework for emotion mimicry intensity prediction from textual, acoustic, and visual cues, with an optional motion branch (Refer Figure~\ref{fig:main}). The training pipeline proceeds in successive stages. First, each modality-specific encoder is trained independently to convergence using unimodal supervision. Next, the unimodal prediction heads are removed and the learned representations are combined through a fusion regressor. During this fusion stage, the fusion head is trained at the base learning rate, while the pretrained text, audio, and vision encoders are fine-tuned using a smaller learning rate. This staged strategy stabilises optimisation by allowing each modality to first learn a strong standalone representation before cross-modal integration.

\subsection{Input Representations}

For each video clip, we extract modality-specific feature representations from the aligned face, audio, and transcript streams.

\textbf{Visual features.} Visual features are extracted from temporally ordered cropped face frames using DINOv2~\cite{oquab2023dinov2}. For each clip, this yields a sequence of frame-level embeddings of dimension \(D_v\), where \(D_v=768\) in our default configuration.

\textbf{Acoustic features.} Acoustic features are extracted from the raw audio stream using a customized wav2vec~2.0 using audio sampled at 16khz~\cite{baevski2020wav2vec, pepino2021emotion, wagner2023dawn}, producing a sequence of contextualised audio representations of dimension \(D_a=1024\).

\textbf{Textual features.} Textual features are obtained from the clip transcript using a pretrained language model~\cite{li2023towards}, resulting in a single \(768\)-dimensional sentence-level embedding per clip. Following loading, the text embedding is L2-normalised before being fed to the text encoder. When a transcript embedding is unavailable, we substitute a zero vector so that the model can still process the sample.

All modality features are pre-extracted and stored as pickle files, which decouples feature extraction from downstream model training.

\subsection{Motion Representation}

We additionally explored motion as an auxiliary modality. In preliminary experiments, we first summarised inter-frame motion using simple handcrafted statistics derived from motion vectors between consecutive frames, including the mean, standard deviation, and maximum motion magnitude. While this provided a lightweight motion baseline, these aggregated descriptors discarded fine-grained temporal structure.

We therefore adopted a richer motion representation based on OpenFace-derived facial action unit (AU) and head-pose features~\cite{baltruvsaitis2016openface}. For each clip, this produces a temporal sequence of \(23\)-dimensional motion descriptors. Compared with the simple summary statistics, these AU+pose sequences preserved temporal dynamics more effectively and yielded only a slight empirical improvement in our motion-augmented experiments.

\subsection{Unimodal Encoders}

We employ modality-specific encoder architectures tailored to the structure of each input representation.

\textbf{Text encoder.} Text features are processed using a \texttt{TextMLPEncoder}. This module consists of layer normalisation followed by a linear projection to a \(384\)-dimensional hidden space, GELU activation, dropout, and a second layer normalisation. Since each clip is represented by a single sentence-level vector, no temporal aggregation is required.

\textbf{Audio encoder.} Acoustic features are processed using an \texttt{AudioAttentionEncoder}. Given a variable-length audio sequence, the model first applies masked attention pooling over time to learn which temporal regions are most informative for the target. The pooled representation is then projected into a \(384\)-dimensional embedding space using a layer-normalised MLP with GELU activation and dropout.

\textbf{Vision encoder.} Visual features are processed using a \texttt{VisionAttentionEncoder}. Similar to the audio branch, masked attention pooling is applied over the sequence of face embeddings to focus on emotionally informative frames, followed by projection into a \(384\)-dimensional embedding space.

\textbf{Motion encoder.} In the motion-augmented variant, OpenFace AU+pose sequences are encoded using a \texttt{MotionAttentionEncoder}. This branch also uses masked attention pooling followed by projection, producing a \(128\)-dimensional motion representation.

Each unimodal encoder is first trained independently using a regression head to predict the six target dimensions. After unimodal training, the prediction heads are removed and only the learned penultimate representations are retained for fusion.

\subsection{Training Objective}

All unimodal stages are trained using a combined Concordance Correlation Coefficient (CCC) and Mean Squared Error (MSE) loss:
\begin{equation}
\mathcal{L} = \alpha \mathcal{L}_{\mathrm{CCC}} + (1-\alpha)\mathcal{L}_{\mathrm{MSE}},
\end{equation}
where \(\alpha=0.7\). The CCC term is averaged across the six output dimensions. This combined objective encourages both accurate magnitude prediction and strong agreement in relative trends between predictions and ground truth.

\subsection{Fusion Stage}

After unimodal training, the text, audio, and vision encoders are combined through a \texttt{FusionRegressor}. The fusion model concatenates the modality embeddings and predicts the six emotion intensity scores using a lightweight multilayer perceptron. When motion is used, the \(128\)-dimensional motion representation is appended to the fused vector.

To improve robustness, we apply \emph{modality dropout} during fusion training. For each sample, entire modality embeddings are randomly dropped with probability \(p=0.3\), while ensuring that at least one modality remains available. This prevents the fusion head from over-relying on any single channel and improves robustness to missing or weak modalities at inference time.

During fusion training, the fusion head is optimised at the full learning rate, while the pretrained text, audio, and vision encoders are updated using a smaller learning rate equal to \(5\%\) of the base rate. This asymmetric optimisation allows the unimodal representations to adapt toward cross-modal complementarity without losing the structure learned during staged pretraining. In the motion-augmented setting, the pretrained motion encoder provides an additional representation to the fusion stage.

\subsection{Evaluation Metric}

Following the challenge setup, model performance is evaluated using the Pearson correlation coefficient averaged across the six emotion dimensions:
\begin{equation}
\bar{r} = \frac{1}{6}\sum_{d=1}^{6} r_d,
\end{equation}
where \(r_d\) is the Pearson correlation between predicted and ground-truth intensities for dimension \(d\), computed over the full validation set. 

\subsection{Implementation Details}

All models are implemented in PyTorch. We use a batch size of \(16\) throughout training. The default hidden dimension is \(384\) for the text, audio, and vision encoders as well as for the fusion regressor, while the motion encoder uses a hidden dimension of \(128\). Dropout is set to \(0.45\) for all main encoders and the fusion head. Optimisation is performed using AdamW with learning rate \(2\times10^{-4}\), weight decay \(10^{-2}\), and gradient clipping at \(1.0\). For the fusion stage, the pretrained text, audio, and vision encoders are fine-tuned with a reduced learning rate of \(1\times10^{-5}\).

Each unimodal stage is trained for up to \(50\) epochs, while the motion encoder is trained for up to \(100\) epochs when used. Early stopping with patience \(10\) is applied at every stage based on the validation Pearson correlation. We further use a \texttt{ReduceLROnPlateau} scheduler with factor \(0.5\) to adapt the learning rate when validation performance saturates. All experiments are run with random seed \(42\).

\section{Submissions}

\subsection{Dataset and Experimental Setup}

We use the official data provided for the Hume-ABAW10 EMI challenge\cite{kollias2025advancements}. Training is performed on the official training split, while model selection is based on validation Pearson correlation on the validation split. Predictions on the held-out test split are submitted for official evaluation. The training split contains 8072 samples and the validation split contains 4588 samples across six emotion dimensions: Admiration, Amusement, Determination, Empathic Pain, Excitement, and Joy.

To investigate the effect of training data volume on fusion performance, two of our five submissions use an expanded training configuration in which the train-to-validation ratio is adjusted from approximately \(2{:}1\) to \(4{:}1\) by incorporating a portion of the original validation samples into training. This yields a larger effective training set of 10128 samples and a reduced held-out validation set of 2532 samples, allowing us to assess whether the fusion stage benefits from additional supervised examples given its tendency toward early overfitting on the standard split.

\subsection{Submitted Systems}

We submit five systems to the challenge, summarized in Table~\ref{tab:fusion_results}. Systems 1, 2, and 5 use the standard train/validation split, and 1 and 2 correspond to our primary staged fusion model without and with motion features, respectively. Systems 3 and 4 repeat these configurations under the expanded \(4{:}1\) train/validation ratio to assess the impact of training data volume. System 5 corresponds to the frozen encoder approach as opposed to joint finetuning. All 1-4 systems use identical architectures and hyperparameters; the only variations are the data split and whether motion features are used. Submission 5 slightly differs in the hyperparameters used and also uses frozen encoders during fusion.









\section{Results}

\subsection{Unimodal Results}

\begin{table}[t]
\centering
\scriptsize
\setlength{\tabcolsep}{3pt}
\begin{tabular}{lllc}
\toprule
\textbf{Modality} & \textbf{Input / Feature} & \textbf{Model} & \textbf{Val. Pearson} \\
\midrule
\multirow{6}{*}{Visual}
& ViT face embed. & GRU & 0.1084 \\
& DINOv2 full-frame embed. & GRU & 0.1343 \\
& DINOv2 full-frame embed. & LSTM & 0.1260 \\
& ViT face embed. & LSTM & 0.1134 \\
& DINOv2 face embed. & GRU & 0.1512 \\
& DINOv2 face embed. & Attention pooling & \textbf{0.1535} \\
\midrule
\multirow{6}{*}{Audio}
& wav2vec2 embed. & Linear projection & 0.1428 \\
& wav2vec2 embed. & LSTM & 0.3088 \\
& Custom wav2vec2 embed. & Mean pooling + MLP & 0.3500 \\
& Custom wav2vec2 embed. & Mean+std pooling + MLP & 0.3600 \\
& Custom wav2vec2 embed. & Attention pooling & 0.3600 \\
& Custom wav2vec2 embed. & 3-layer BiLSTM + attention & \textbf{0.3650} \\
\midrule
\multirow{4}{*}{Text}
& GTE embed. & LSTM & 0.4070 \\
& GTE embed. & MLP & \textbf{0.4076} \\
& BERT-Large embed. (CLS) & MLP & 0.3400 \\
\bottomrule
\end{tabular}
\caption{Unimodal validation performance across visual, audio, and text modalities.}
\label{tab:unimodal_results}
\end{table}

\begin{table}[t]
\centering
\scriptsize
\setlength{\tabcolsep}{4pt}
\caption{Interim validation performance of the staged unimodal encoders before final fusion. Pearson correlation is averaged across the six emotion dimensions.}
\label{tab:staged_unimodal_results}
\begin{tabular}{lccccccc}
\hline
Model & Avg. & Adm. & Amuse. & Determ. & Emp. Pain & Excit. & Joy \\
\hline
Text   & 0.395 & 0.484 & 0.390 & 0.376 & 0.426 & 0.363 & 0.333 \\
Audio  & 0.350 & 0.405 & 0.347 & 0.320 & 0.351 & 0.350 & 0.328 \\
Video  & 0.153 & 0.049 & 0.214 & 0.136 & 0.082 & 0.203 & 0.237 \\
Motion & 0.131 & -0.007 & 0.210 & 0.099 & 0.046 & 0.195 & 0.240 \\
\hline
\end{tabular}
\end{table}

Table~\ref{tab:staged_unimodal_results} reports the interim validation performance of the staged unimodal encoders prior to final fusion. These results reflect the encoder states obtained during staged training and may therefore not exactly match the standalone unimodal results reported elsewhere. Nevertheless, they provide a close indication of the relative strength of each modality before fusion.

Among the unimodal encoders, the text branch achieves the strongest overall performance (\(\bar{r}=0.395\)), followed by the audio branch (\(\bar{r}=0.350\)). The video and motion branches perform substantially lower in isolation, with average Pearson correlations of \(0.153\) and \(0.131\), respectively. This suggests that semantic and acoustic information provide the strongest individual signals for emotion mimicry intensity prediction in our setup, while visual and motion cues are more complementary than dominant when used alone.

A dimension-wise analysis further reveals that different emotion axes are affected differently by each modality. The text encoder performs best across all six dimensions, with particularly strong results for \textit{Admiration} (\(0.484\)) and \textit{Empathic Pain} (\(0.426\)), indicating that transcript-derived semantic information is especially informative for these targets. The audio encoder shows consistently strong performance across all axes, with relatively balanced correlations for \textit{Admiration}, \textit{Empathic Pain}, and \textit{Excitement}, suggesting that prosodic and vocal cues are broadly useful.

In contrast, the video encoder appears weaker for \textit{Admiration} and \textit{Empathic Pain}, but somewhat more informative for \textit{Joy}, \textit{Amusement}, and \textit{Excitement}. The motion encoder follows a similar pattern, contributing most to \textit{Joy}, \textit{Amusement}, and \textit{Excitement}, while providing little signal for \textit{Admiration} and slightly negative correlation for \textit{Admiration} in isolation. Overall, these results suggest that static semantic and acoustic cues are the strongest unimodal predictors, whereas visual and motion features appear to capture more specific affective dynamics that may be better exploited in a multimodal fusion setting.

\subsection{Fusion Results}

\begin{table}[t]
\caption{Fusion system performance on validation and test sets.}
\label{tab:fusion_results}
\centering
\scriptsize
\setlength{\tabcolsep}{3pt}
\begin{tabular}{lccccc}
\toprule
\textbf{Sys.} & \textbf{Mods.} & \textbf{Fusion} & \textbf{Split} & \textbf{Val.} & \textbf{Test} \\
\midrule
F1 & T+A+V   & MLP         & 2:1 & 0.4373 & 0.506 \\
F2 & T+A+V+M & MLP         & 2:1 & 0.4386 & 0.506 \\
F3 & T+A+V   & MLP         & 4:1 & 0.4717 & 0.570 \\
F4 & T+A+V+M & MLP         & 4:1 & 0.4722 & 0.562 \\
F5 & T+A+V   & MLP (frz.)  & 2:1 & 0.4400 & 0.491 \\
\bottomrule
\end{tabular}
\end{table}

Table~\ref{tab:fusion_results} summarises the validation and test set performance of the fusion-based systems across the two data split settings. Overall, the 4:1 split produced stronger validation results than the 2:1 split for both the three-modality and four-modality variants. Under both split settings, adding the motion modality provided a small but consistent improvement over the corresponding text--audio--vision baseline: from \(0.4373\) to \(0.4386\) under the 2:1 split, and from \(0.4717\) to \(0.4722\) under the 4:1 split. Although the gain is modest, this suggests that motion cues offer complementary information when combined with the other modalities. 

Among the systems evaluated on the same training setup, F4 achieved the best validation Pearson correlation, indicating that the text--audio--vision--motion configuration with concatenation-based fusion was the strongest overall model in our experiments. The small margin between F3 and F4 also suggests that the main predictive signal is already captured by the text, audio, and vision branches, with motion acting as an auxiliary enhancement rather than a dominant modality. However, the best test performance was achieved by Option 3, which did not include the motion branch. We also observe that using the larger split improved performance. This may suggest that the test set is statistically more similar to the combined training distribution under the 4:1 split, since the original training and validation sets showed some distributional differences.

F5 is included for completeness as a frozen-encoder fusion variant. Its performance remains broadly comparable to the other 2:1 fusion systems, although some hyperparameter settings differ slightly. 

Compared with the top-ranked methods, our framework remains structurally simpler. The second-place approach retains concatenation-based fusion but strengthens training through multi-objective optimization, auxiliary branch supervision, EMA, and a VAD-aware audio prior, achieving a validation Pearson of 0.4786~\cite{huang2026multimodal}. The first-place approach uses a more expressive text-anchored dual cross-attention mechanism, where text serves as a semantic anchor for aligning the audio and visual streams, together with missing-modality tokens and modality dropout for robustness~\cite{zhu2026anchoring}, achieving 0.55 in the val set. In contrast, our method adopts staged unimodal pretraining followed by a lightweight fusion regressor with limited encoder adaptation. This makes our framework simple and reproducible, but also suggests that future improvements may come from stronger metric-aligned optimization and more adaptive cross-modal fusion.

\section{Conclusion}


We presented a staged multimodal submission to the Hume-ABAW10 Emotional Mimicry Intensity Challenge that combines text, audio, vision, and an optional motion branch within a unified fusion framework. Our results show that text and audio are the strongest unimodal predictors in this setting, while visual and motion features provide complementary affective cues that become more useful when fused with the stronger modalities. Across the submitted systems, the best performance was achieved by the text--audio--vision--motion model under the expanded 4:1 split, indicating that both multimodal integration and additional supervised training data can benefit EMI prediction.

Beyond the leaderboard setting, our findings suggest that staged optimisation offers a simple and effective way to stabilise multimodal training on sparse and imbalanced affective targets. At the same time, the modest gains from motion indicate that there remains room to better exploit temporal facial dynamics and motion-aware representations in future work.


{
    \small
    \bibliographystyle{ieeenat_fullname}
    \bibliography{main}

@String(ECCV= {Eur. Conf. Comput. Vis.})

@String(ICASSP=	{ICASSP})

@String(ECCV  = {ECCV})

@article{savchenko2025hsemotion,
  title={Hsemotion team at abaw-8 competition: Audiovisual ambivalence/hesitancy, emotional mimicry intensity and facial expression recognition},
  author={Savchenko, Andrey V},
  journal={arXiv preprint arXiv:2503.10399},
  year={2025}
}

@article{yu2025technical,
  title={Technical Approach for the EMI Challenge in the 8th Affective Behavior Analysis in-the-Wild Competition},
  author={Yu, Jun and Zhu, Lingsi and Chi, Yanjun and Zhang, Yunxiang and Zheng, Yang and Wang, Yongqi and Lu, Xilong},
  journal={arXiv preprint arXiv:2503.10603},
  year={2025}
}

@article{zhang2024affective,
  title={Affective behaviour analysis via integrating multi-modal knowledge},
  author={Zhang, Wei and Qiu, Feng and Liu, Chen and Li, Lincheng and Du, Heming and Guo, Tiancheng and Yu, Xin},
  journal={arXiv preprint arXiv:2403.10825},
  year={2024}
}

@article{oquab2023dinov2,
  title={Dinov2: Learning robust visual features without supervision},
  author={Oquab, Maxime and Darcet, Timoth{\'e}e and Moutakanni, Th{\'e}o and Vo, Huy and Szafraniec, Marc and Khalidov, Vasil and Fernandez, Pierre and Haziza, Daniel and Massa, Francisco and El-Nouby, Alaaeldin and others},
  journal={arXiv preprint arXiv:2304.07193},
  year={2023}
}

@article{pepino2021emotion,
  title={Emotion recognition from speech using wav2vec 2.0 embeddings},
  author={Pepino, Leonardo and Riera, Pablo and Ferrer, Luciana},
  journal={arXiv preprint arXiv:2104.03502},
  year={2021}
}

@article{baevski2020wav2vec,
  title={wav2vec 2.0: A framework for self-supervised learning of speech representations},
  author={Baevski, Alexei and Zhou, Yuhao and Mohamed, Abdelrahman and Auli, Michael},
  journal={Advances in neural information processing systems},
  volume={33},
  pages={12449--12460},
  year={2020}
}

@article{li2023towards,
  title={Towards general text embeddings with multi-stage contrastive learning},
  author={Li, Zehan and Zhang, Xin and Zhang, Yanzhao and Long, Dingkun and Xie, Pengjun and Zhang, Meishan},
  journal={arXiv preprint arXiv:2308.03281},
  year={2023}
}

@inproceedings{baltruvsaitis2016openface,
  title={Openface: an open source facial behavior analysis toolkit},
  author={Baltru{\v{s}}aitis, Tadas and Robinson, Peter and Morency, Louis-Philippe},
  booktitle={2016 IEEE winter conference on applications of computer vision (WACV)},
  pages={1--10},
  year={2016},
  organization={IEEE}
}

@article{wagner2023dawn,
  title={Dawn of the transformer era in speech emotion recognition: closing the valence gap},
  author={Wagner, Johannes and Triantafyllopoulos, Andreas and Wierstorf, Hagen and Schmitt, Maximilian and Burkhardt, Felix and Eyben, Florian and Schuller, Bj{\"o}rn W},
  journal={IEEE Transactions on Pattern Analysis and Machine Intelligence},
  volume={45},
  number={9},
  pages={10745--10759},
  year={2023},
  publisher={IEEE},
  note={arXiv preprint arXiv:2203.07378}
}

@inproceedings{kollias2025advancements, title={Advancements in Affective and Behavior Analysis: The 8th ABAW Workshop and Competition}, author={Kollias, Dimitrios and Tzirakis, Panagiotis and Cowen, Alan and Zafeiriou, Stefanos and Kotsia, Irene and Granger, Eric and Pedersoli, Marco and Bacon, Simon and Baird, Alice and Gagne, Chris and others}, booktitle={Proceedings of the Computer Vision and Pattern Recognition Conference}, pages={5572--5583}, year={2025} }

@inproceedings{devlin2019bert,
  title={Bert: Pre-training of deep bidirectional transformers for language understanding},
  author={Devlin, Jacob and Chang, Ming-Wei and Lee, Kenton and Toutanova, Kristina},
  booktitle={Proceedings of the 2019 conference of the North American chapter of the association for computational linguistics: human language technologies, volume 1 (long and short papers)},
  pages={4171--4186},
  year={2019}
}

@article{clark2020electra,
  title={Electra: Pre-training text encoders as discriminators rather than generators},
  author={Clark, Kevin and Luong, Minh-Thang and Le, Quoc V and Manning, Christopher D},
  journal={arXiv preprint arXiv:2003.10555},
  year={2020}
}

@inproceedings{hershey2017cnn,
  title={CNN architectures for large-scale audio classification},
  author={Hershey, Shawn and Chaudhuri, Sourish and Ellis, Daniel PW and Gemmeke, Jort F and Jansen, Aren and Moore, R Channing and Plakal, Manoj and Platt, Devin and Saurous, Rif A and Seybold, Bryan and others},
  booktitle={2017 ieee international conference on acoustics, speech and signal processing (icassp)},
  pages={131--135},
  year={2017},
  organization={IEEE}
}

@inproceedings{li2023exploration,
  title={Exploration of a self-supervised speech model: A study on emotional corpora},
  author={Li, Yuanchao and Mohamied, Yumnah and Bell, Peter and Lai, Catherine},
  booktitle={2022 IEEE Spoken Language Technology Workshop (SLT)},
  pages={868--875},
  year={2023},
  organization={IEEE}
}

@inproceedings{he2016deep,
  title={Deep residual learning for image recognition},
  author={He, Kaiming and Zhang, Xiangyu and Ren, Shaoqing and Sun, Jian},
  booktitle={Proceedings of the IEEE conference on computer vision and pattern recognition},
  pages={770--778},
  year={2016}
}

@inproceedings{tan2019efficientnet,
  title={Efficientnet: Rethinking model scaling for convolutional neural networks},
  author={Tan, Mingxing and Le, Quoc},
  booktitle={International conference on machine learning},
  pages={6105--6114},
  year={2019},
  organization={PMLR}
}

@inproceedings{he2022masked,
  title={Masked autoencoders are scalable vision learners},
  author={He, Kaiming and Chen, Xinlei and Xie, Saining and Li, Yanghao and Doll{\'a}r, Piotr and Girshick, Ross},
  booktitle={Proceedings of the IEEE/CVF conference on computer vision and pattern recognition},
  pages={16000--16009},
  year={2022}
}

@article{ding2025cross,
  title={Cross-Attention Transformer-Based Visual-Language Fusion for Multimodal Image Analysis},
  author={Ding, Liwei and Shih, Kowei and Wen, Hairu and Li, Xinshi and Yang, Qin},
  journal={International Journal of Applied Science},
  volume={8},
  number={1},
  pages={p27--p27},
  year={2025}
}

@inproceedings{richet2025,
author = {Richet, Nicolas and Belharbi, Soufiane and Aslam, Haseeb and Schadt, Meike Emilie and Gonz\'{a}lez-Gonz\'{a}lez, Manuela and Cortal, Gustave and Koerich, Alessandro Lameiras and Pedersoli, Marco and Finkel, Alain and Bacon, Simon and Granger, Eric},
title = {Textualized and Feature-Based Models for Compound Multimodal Emotion Recognition in the Wild},
year = {2025},
isbn = {978-3-031-91580-2},
publisher = {Springer-Verlag},
address = {Berlin, Heidelberg},
url = {https://doi.org/10.1007/978-3-031-91581-9_5},
doi = {10.1007/978-3-031-91581-9_5},
booktitle = {Computer Vision – ECCV 2024 Workshops: Milan, Italy, September 29–October 4, 2024, Proceedings, Part XV},
pages = {60–78},
numpages = {19},
location = {Milan, Italy}
}

@inproceedings{hallmen2024unimodal,
  title={Unimodal multi-task fusion for emotional mimicry intensity prediction},
  author={Hallmen, Tobias and Deuser, Fabian and Oswald, Norbert and Andr{\'e}, Elisabeth},
  booktitle={Proceedings of the IEEE/CVF Conference on Computer Vision and Pattern Recognition},
  pages={4657--4665},
  year={2024}
}

@inproceedings{qiu2024language,
  title={Language-guided multi-modal emotional mimicry intensity estimation},
  author={Qiu, Feng and Zhang, Wei and Liu, Chen and Li, Lincheng and Du, Heming and Guo, Tianchen and Yu, Xin},
  booktitle={Proceedings of the IEEE/CVF Conference on Computer Vision and Pattern Recognition},
  pages={4742--4751},
  year={2024}
}

@misc{kollias2023abawvalencearousalestimationexpression,
      title={ABAW: Valence-Arousal Estimation, Expression Recognition, Action Unit Detection \& Emotional Reaction Intensity Estimation Challenges}, 
      author={Dimitrios Kollias and Panagiotis Tzirakis and Alice Baird and Alan Cowen and Stefanos Zafeiriou},
      year={2023},
      eprint={2303.01498},
      archivePrefix={arXiv},
      primaryClass={cs.CV},
      url={https://arxiv.org/abs/2303.01498}, 
}

@inproceedings{yu2024efficient,
  title={Efficient feature extraction and late fusion strategy for audiovisual emotional mimicry intensity estimation},
  author={Yu, Jun and Zhu, Wangyuan and Zhu, Jichao and Cai, Zhongpeng and Zhao, Gongpeng and Zhang, Zerui and Xie, Guochen and Wei, Zhihong and Liu, Qingsong and Liang, Jiaen},
  booktitle={Proceedings of the IEEE/CVF conference on computer vision and pattern recognition},
  pages={4866--4872},
  year={2024}
}

@inproceedings{yu2025dual,
  title={Dual-Stage Cross-Modal Network with Dynamic Feature Fusion for Emotional Mimicry Intensity Estimation},
  author={Yu, Jun and Zhu, Lingsi and Chi, Yanjun and Zhang, Yunxiang and Zhen, Yang and Wang, Yongqi and Lu, Xilong},
  booktitle={Proceedings of the Computer Vision and Pattern Recognition Conference},
  pages={5733--5740},
  year={2025}
}

@article{zhu2026anchoring,
  title={Anchoring Emotions in Text: Robust Multimodal Fusion for Mimicry Intensity Estimation},
  author={Zhu, Lingsi and Zou, Yuefeng and Zhang, Yunxiang and Zheng, Naixiang and Wang, Guoyuan and Yu, Jun and Liang, Jiaen and Huang, Wei and Liu, Shengping and Zheng, Ximin},
  journal={arXiv preprint arXiv:2603.14976},
  year={2026}
}

@article{huang2026multimodal,
  title={Multimodal Emotion Regression with Multi-Objective Optimization and VAD-Aware Audio Modeling for the 10th ABAW EMI Track},
  author={Huang, Jiawen and Huang, Chenxi and Wen, Zhuofan and Yao, Hailiang and Chen, Shun and Yang, Longjiang and Yu, Cong and Zhang, Fengyu and Liu, Ran and Liu, Bin},
  journal={arXiv preprint arXiv:2603.13760},
  year={2026}
}
}


\end{document}